\documentclass[letterpaper]{article} 
\usepackage{aaai25}  
\usepackage{times}  
\usepackage{helvet}  
\usepackage{courier}  
\usepackage[hyphens]{url}  
\usepackage{graphicx} 
\usepackage{natbib}  
\usepackage{caption} 
\usepackage{algorithm}
\usepackage{algorithmic}

\usepackage{amsmath}
\usepackage{multirow}
\usepackage{amssymb}
\usepackage{xcolor}

\usepackage{newfloat}
\usepackage{listings}
\DeclareCaptionStyle{ruled}{labelfont=normalfont,labelsep=colon,strut=off} 
\floatstyle{ruled}
\newfloat{listing}{tb}{lst}{}
\floatname{listing}{Listing}
\title{Embedding and Enriching Explicit Semantics for Visible-Infrared Person Re-Identification}
\author{
   Neng~Dong, Shuanglin~Yan, Liyan~Zhang, Jinhui~Tang
}
\affiliations{
}

\usepackage{bibentry}

\begin{document}

\maketitle

\begin{abstract}
Visible-infrared person re-identification (VIReID) retrieves pedestrian images with the same identity across different modalities. Existing methods learn visual content solely from images, lacking the capability to sense high-level semantics. In this paper, we propose an Embedding and Enriching Explicit Semantics (EEES) framework to learn semantically rich cross-modality pedestrian representations. Our method offers several contributions. First, with the collaboration of multiple large language-vision models, we develop Explicit Semantics Embedding (ESE), which automatically supplements language descriptions for pedestrians and aligns image-text pairs into a common space, thereby learning visual content associated with explicit semantics. Second, recognizing the complementarity of multi-view information, we present Cross-View Semantics Compensation (CVSC), which constructs multi-view image-text pair representations, establishes their many-to-many matching, and propagates knowledge to single-view representations, thus compensating visual content with its missing cross-view semantics. Third, to eliminate noisy semantics such as conflicting color attributes in different modalities, we design Cross-Modality Semantics Purification (CMSP), which constrains the distance between inter-modality image-text pair representations to be close to that between intra-modality image-text pair representations, further enhancing the modality-invariance of visual content. Finally, experimental results demonstrate the effectiveness and superiority of the proposed EEES.
\end{abstract}

%

\section{Introduction}

Person re-identification (ReID) aims to match images depicting the same individual across cameras, a critical component of intelligent security with profound research implications. Despite significant advancements \cite{agw, dc-former, entdnet}, most existing algorithms focus on single-modality retrieval, neglecting the requirements of round-the-clock surveillance systems where infrared images dominate nighttime scenarios. To address this challenge, visible-infrared person ReID (VIReID) has emerged to retrieve visible images corresponding to the identity of a given infrared query, and vice versa \cite{vireid}.

\begin{figure}[t!]
	\centering
        \includegraphics[width=3.3in,height=1.8in]{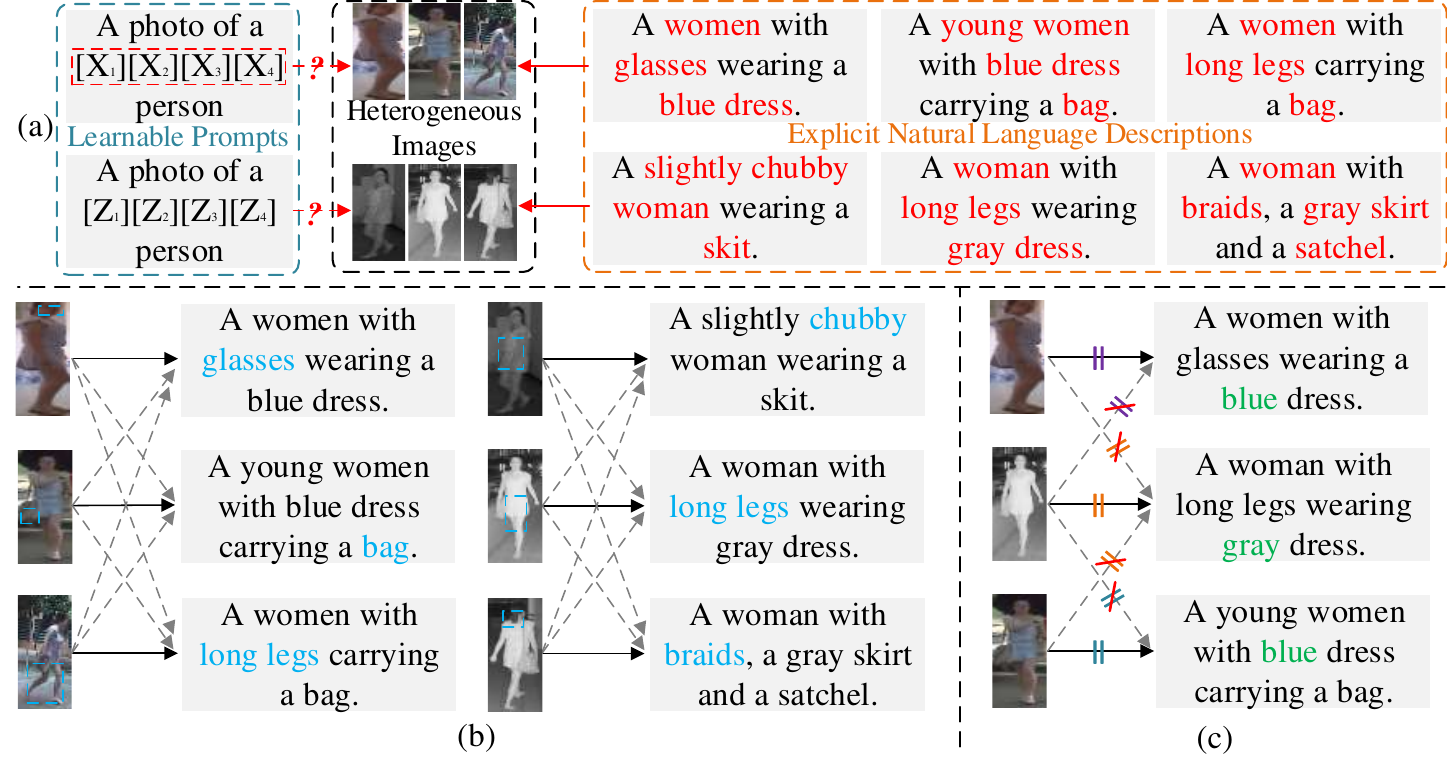}
	\caption{The core motivation of our EEES framework arises from several key observations: (a) Language descriptions produced by off-the-shelf large language-vision generation models surpass learnable prompts in clarity and detail. (b) Multi-view images/texts exhibit significant complementary attributes. (c) Noise information such as color clues leads to semantic conflicts between paired cross-modality images.}
	\label{Fig:1}
\end{figure}

VIReID focuses on aligning the feature distribution of heterogeneous images, addressing this challenge with two distinct approaches. One approach involves the generative-based method \cite{aligngan, hicmd}, which attempts to bridge the modality gap through style transfer technology. However, noise introduced during generation compromises feature discriminability. The alternative approach, generative-free method \cite{mid, caj+}, emphasizes network design and metric function optimization. Comparatively, the generative-free method has demonstrated superior effectiveness in aligning modalities and currently stands as the predominant solution. Nevertheless, addressing VIReID solely through a vision-centric approach is suboptimal, as visual content learned from images alone fails to capture semantic information. The advent of large language-vision matching (LLVM) \cite{clip} provides a promising solution to this limitation. Recent research \cite{csdn} indicates that there is no modality discrepancy in language descriptions corresponding to heterogeneous images, making them well-suited for aligning cross-modality visual feature distribution.

As is well-known, pedestrian images typically lack accompanying language descriptions. An effective approach to address this issue is designing learnable prompts for images \cite{clipreid}, as illustrated in Figure \ref{Fig:1}(a). Although feasible, this strategy encounters several challenges: 1) Uncertainty. The set trainable words are unknown, raising questions about what the semantic information they represent; 2) Coarseness. Typically, pedestrian images with the same identity share a common prompt, and only four learnable tokens are allocated for identity depiction, which is insufficient for the cross-view and fine-grained nature of VIReID; 3) Cumbersomeness. Rather than end-to-end, the paradigm of learnable prompts requires a meticulously designed two-stage training process. Recently, significant advancements in large language-vision generation (LLVG) \cite{blip2, llava} have demonstrated a potent ability to generate clear and detailed image descriptions. This inspires a solution to the aforementioned challenges: automatically supplementing textual data to acquire explicit semantics of pedestrians and embedding them into visual representations via image-text matching. Notably, the general matching strategy only considers the one-to-one correspondence between a single image and its paired text. However, as depicted in Figure \ref{Fig:1}(b), cross-view images sharing the same identity exhibit diverse visual cues, accompanied by varying semantics in their paired descriptions. Consequently, the one-to-one matching strategy may impede the learning of comprehensive knowledge as it ignores the rich complementary information inherent in multi-view images and texts. Additionally, as shown in Figure \ref{Fig:1}(c), language descriptions generated for heterogeneous images often exhibit conflicting semantics (e.g., color attributes), potentially undermining the modality-invariance of visual content. Therefore, it is necessary to eliminate such noise information during semantics embedding.

In this paper, we present a novel framework named \textbf{E}mbedding and \textbf{E}nriching \textbf{E}xplicit \textbf{S}emantics (\textbf{EEES}), aimed at learning pedestrian visual representations associated with rich high-quality semantics to mitigate the modality gap in VIReID. The framework consists of three main modules: Explicit Semantics Embedding (ESE), Cross-View Semantics Compensation (CVSC), and Cross-Modality Semantics Purification (CMSP). Specifically, ESE employs an off-the-shelf image-text generation model to automatically supplement language descriptions for pedestrians and uses contrastive learning to align image-text pair representations into a common space, embedding explicit semantics into cross-modality visual contents. CVSC fuses image (text) features sharing the same identity across different views to construct multi-view image-text pair representations, and establishes the correspondence between them, thereby learning semantically rich visual contents. Since only single-view images are available during inference, CVSC propagates information from multi-view representations to single-view ones through knowledge distillation, compensating visual contents with their missing cross-view semantics. CMSP constrains the distance between inter-modality image-text pair representations to be close to that of intra-modality-modality image-text pair representations, avoiding the embedding of conflicting semantics. The proposed EEES is trained end-to-end, with only the visual side used to extract single-view representations for testing.

Our main contributions are summarized as follows: 

\begin{itemize}
\item We propose a novel EEES framework to embed rich explicit semantics into cross-modality visual representations. To the best of our knowledge, we are the first to explore the collaboration of multiple language-vision models to mitigate the modality discrepancy in VIReID.
\item We propose CVSC, which mines many-to-many image-text correspondences to compensate visual representations with their missing cross-view semantics, and CMSP, which eliminates noisy semantics to strengthen the modality-invariance of visual representations.
\item Extensive experiments across two benchmark datasets demonstrate that EEES achieves new state-of-the-art performance, with each component contributing effectively.
\end{itemize}

\begin{figure*}[t!]
	\centering
	\includegraphics[width=7.0in,height=3.2in]{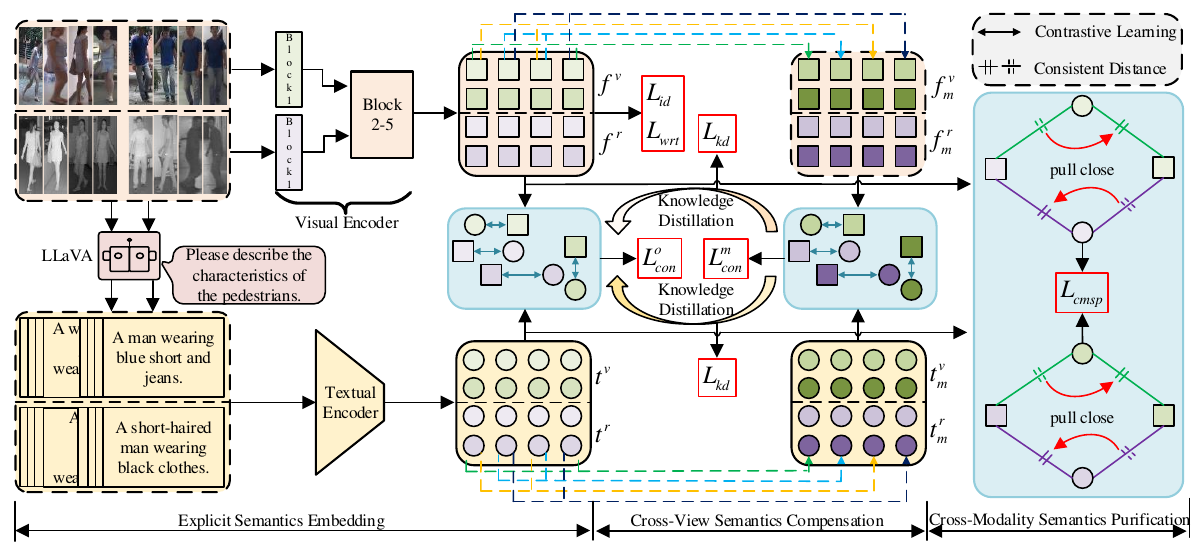}
	\caption{Overview of our EEES. It comprises ESE, CVSC, and CMSP. ESC supplements language descriptions for images and aligns image-text pairs into a common space. CVSC fuses image (text) features with the same identity across different views, establishes correspondences between multi-view image-text pair representations, and transfers knowledge from multi-view representations to single-view ones. CMSP constrains the distance between inter-modality image-text pair representations to be close to that of intra-modality image-text pair representations. During inference, only the visual side is used.}
	\label{Fig:2}
\end{figure*}

\section{Related Work}
\subsection{Visible-Infrared Person Re-Identification}
VIReID is a challenging task due to the significant modality gap between visible and infrared images. One intuitive approach is to transfer images from one modality to the style of another or generate intermediate images containing information from both modalities. For instance, JSIA \cite{jsia} employed feature decoupling and cycle generation to produce high-quality cross-modality paired images. Given the substantial gap between heterogeneous data hinders style transfer, XIV-ReID \cite{x-modality} introduced an auxiliary X-modality to reconcile the infrared and visible modalities. To prevent identity information loss during generation, GC-IFS \cite{gcifs} designed a cross-modality contrastive loss to ensure the generated images retain a consistent identity with the original ones. Although generative-based methods are intuitive and effective, they are prone to model collapse and susceptible to introducing noise.

The generative-free method has recently garnered increased attention as it circumvents the limitations of generative approaches. This method primarily focuses on aligning cross-modality features by constructing appropriate networks or metric functions. For instance, Zero-Padding \cite{vireid} evaluated the suitability of four networks for VIReID and proposed a one-stream structure with a zero-padding strategy. AGW \cite{agw} devised a weighted regularization triplet loss to optimize the relative distance between positive and negative pairs in both intra-modality and inter-modality. DEEN \cite{deen} designed an embedding expansion network containing multiple dilated convolutional blocks to enhance feature diversity. To capture fine-grained information, DMA \cite{dma} aligned heterogeneous features at the local level. However, current methodologies treat VIReID as a vision-only task, resulting in the visual content lacking high-level semantic information. Although a recent study \cite{csdn} addressed this issue using LLVM, the introduced learnable prompts were found to compromise semantics quality. In this study, we explore a multi-model collaborative paradigm to address this challenge.

\subsection{Large Language-Vision Pre-training}
Large language-vision pre-training has become a significant research focus, unifying computer vision and natural language processing while demonstrating remarkable performance in the fields of LLVM and LLVG. Contrastive Language-Image Pre-training (CLIP) \cite{clip}, a prominent LLVM model, excels at embedding high-level semantics into visual content by bridging the connection between image-text pairs, thereby advancing various downstream visual tasks \cite{cris,cfine}. In the field of ReID, CLIP-ReID \cite{clipreid} introduced a learnable prompt to acquire the implicit semantics of pedestrians. In the realm of VIReID, CSDN \cite{csdn} confirmed that there is no modality gap in language descriptions corresponding to heterogeneous images, allowing CLIP to naturally align visible and infrared visual representations. However, the semantics represented by learned prompts are unknown and coarse. Additionally, CSDN fails to consider the complementarity of multi-view information, and conflicting attributes exist in generated bi-modality language descriptions. This insight inspires us to further develop CVSC and CMSP to enrich and purify semantics.

\section{Methodology}
\subsection{Preliminaries}
Formally, we define the visible and infrared image sets as $\{x^{v}_i\}_{i=1}^{N_v}$ and $\{x^{r}_i\}_{i=1}^{N_r}$, where $N_v$ and $N_r$ represent the sizes of these two heterogeneous data, respectively. The label set is denoted as $\{y_i\}_{i=1}^{N_p}$, with $N_p$ indicates the number of identities. In each mini-batch, $N$ paired cross-modality images $\{x^{v}_i, x^{r}_i\}_{i=1}^N$ are randomly sampled and their visual representations $\{f^{v}_i, f^{r}_i\}_{i=1}^N \in R^{N\times d}$ are extracted. We employ identity loss and weighted regularization triplet loss to optimize the network, with the former can be formulated as:
\begin{equation}
    L_{id} =-\frac{1}{N}\sum_{i=1}^{N}q_{i}\log(p_{i}^{v})-\frac{1}{N}\sum_{i=1}^{N}q_{i}\log(p_{i}^{r}),
\end{equation}
here $q_{i}$ is the one-hot vector of identity label $y_{i}$, and $p_{i}^{v}$ and $p_{i}^{r}$ represent classification results of $f_{i}^{v}$ and $f_{i}^{r}$, respectively.

The weighted regularization triplet loss aims to bring cross-modality positive sample pairs closer while pushing negative ones apart. For convenience, here we denote the visual representations as $\{f_i\}_{i=1}^{2N} = \{(f_i^{v},f_i^{r})\}_{i=1}^{N}$ and formulate this loss as:
\begin{equation}
    L_{wrt}=\frac{1}{2N}\sum_{i=1}^{2N}\log(1+\exp(\sum\nolimits_{ij}d^{wp}_{ij}-\sum\nolimits_{ik}d^{wn}_{ik})),
\end{equation}
where $j$ and $k$ are indices of the positive and negative representations corresponding to $f_{i}$; $d^{wp}_{ij}$ and $d^{wn}_{ij}$ denote the weighted Euclidean distances of positive and negative pairs.

\subsection{Emdedding and Enriching Explicit Semantics}

Most existing frameworks treat VIReID as a purely visual task, lacking the ability to capture semantics associated with visual content. Although CSDN introduces CLIP to address this limitation, the uncertainty and coarseness of the learned implicit semantics hinder performance improvement. Additionally, acquiring richer and purer semantics can further alleviate the modality gap. In this paper, we propose an Embedding and Enriching Explicit Semantics (EEES) framework, which includes Explicit Semantics Embedding (ESE), Cross-View Semantics Compensation (CVSC), and Cross-Modality Semantics Purification (CMSP) to facilitate the learning of representations with high-quality explicit semantics. These components are detailed below.

\subsubsection{Explicit Semantics Embedding} Our ESE involves two processes: supplementing language descriptions with LLVG and aligning image-text pairs with LLVM.

\noindent (1) With the assistance of LLaVA \cite{llava}, an advanced LLVG model, we supplement language descriptions corresponding to pedestrian images. As illustrated in Figure \ref{Fig:2}, given a pedestrian image, we send the request command 'Please describe the characteristics of the pedestrian image' to LLaVa. It responds with a natural language description 'A short-haired man wearing black clothes'. This description provides clearer and more detailed explicit semantics, such as gender, hairstyle, and clothing, compared to the learnable prompt 'A photo of a [X$_{1}$][X$_{2}$][X$_{3}$][X$_{4}$] person' in CSDN. Notably, LLaVA operates without the need for training.

\noindent (2) Suppose the generated cross-modality language bases are $\{l^{v}_i\}_{i=1}^{N_v}$ and $\{l^{r}_i\}_{i=1}^{N_r}$. In each mini-batch, we sample $\{l^{v}_i, l^{r}_i\}_{i=1}^N$ corresponding to $\{x^{v}_i, x^{r}_i\}_{i=1}^N$ and input them into the textual encoder of CLIP to extract representations $\{t^{v}_i, t^{r}_i\}_{i=1}^N \in R^{N\times d}$. To associate the semantic information in $\{t^{v}_i, t^{r}_i\}_{i=1}^N$ with $\{f^{v}_i, f^{r}_i\}_{i=1}^N$, we employ contrastive loss \cite{contrastivelearning} to align them into a common space:
\begin{equation}
\label{Eq3}
\begin{array}{ll}
     L_{con}^{o}=L_{i2t}^{o}+L_{t2i}^{o},
\end{array}
\end{equation}
where
\begin{equation}
\label{Eq4}
\begin{aligned}
     L_{i2t}^{o}=&-\frac{1}{N}\sum_{i=1}^{N}\log\frac{\exp(s(f_{i}^{v},t_{i}^{v}))}{\sum\nolimits_{j=1}^{N}\exp(s(f_{i}^{v},t_{j}^{v}))} \\
     &-\frac{1}{N}\sum_{i=1}^{N}\log\frac{\exp(s(f_{i}^{r},t_{i}^{r}))}{\sum\nolimits_{j=1}^{N}\exp(s(f_{i}^{r},t_{j}^{r}))},
\end{aligned}
\end{equation}
\begin{equation}
\label{Eq5}
\begin{aligned}
L_{t2i}^{o}=&-\frac{1}{N}\sum_{i=1}^{N}\log\frac{\exp(s(t_{i}^{v},f_{i}^{v}))}{\sum\nolimits_{j=1}^{N}\exp(s(t_{i}^{v},f_{j}^{v}))} \\
     &-\frac{1}{N}\sum_{i=1}^{N}\log\frac{\exp(s(t_{i}^{r},f_{i}^{r}))}{\sum\nolimits_{j=1}^{N}\exp(s(t_{i}^{r},f_{j}^{r}))},
\end{aligned}
\end{equation}
where $s(\cdot)$ represents the similarity between two vectors. $L_{i2t}^{o}$ and $L_{t2i}^{o}$ denote the alignment of image-to-text and text-to-image, respectively. This process enables the model to sense explicit pedestrian semantics. However, it only considers the one-to-one matching between image and text, neglecting the complementarity of cross-view information. This limitation motivates our proposed CVSC as below.

\subsubsection{Cross-View Semantics Compensation} Our CVSC involves three processes: constructing multi-view representations, establishing many-to-many correspondences, and propagating information to single-view representations.

\noindent(1) Multiple images of the same pedestrian from different views reveal diverse identity clues, providing more comprehensive discriminative information than single-view images. Likewise, multiple descriptions corresponding to these images offer richer semantics than a single one. To this end, We construct cross-modality multi-view visual and textual representations, $\{(f_{m,i}^{v}, f_{m,i}^{r})\}_{i=1}^{N}$ and $\{(t_{m,i}^{v}, t_{m,i}^{r})\}_{i=1}^{N}$, to integrate cross-view information into the current view. Taking $f_{i}^{v}$ and $t_{i}^{v}$ as examples, we randomly select $M$ visual and textual features sharing the same identity as $f_{i}^{v}$ and $t_{i}^{v}$ and fuse them respectively by sum averaging:
\begin{equation}
    f_{m,i}^{v} =\frac{1}{M+1}(f_{i}^{v}+\sum_{m=1}^{M}f_{m}^{v}),
\end{equation}
\begin{equation}
    t_{m,i}^{v} =\frac{1}{M+1}(t_{i}^{v}+\sum_{m=1}^{M}t_{m}^{v}),
\end{equation}
here $M$ indicates the number of cross-view representations. Similarly, $f_{m,i}^{r}$ and $t_{m,i}^{r}$ can be obtained in the same manner.

\noindent(2) We apply contrastive losses, similar to Eqs. \ref{Eq3}, \ref{Eq4} and \ref{Eq5}, on $\{(f_{m,i}^{v}, t_{m,i}^{v})\}_{i=1}^{N}$ and $\{(f_{m,i}^{r}, t_{m,i}^{r})\}_{i=1}^{N}$ to achieve many-to-many image-text matching, thereby learning comprehensive visual representations associated with rich semantics:
\begin{equation}
     L_{con}^{m}=L_{i2t}^{m}+L_{t2i}^{m},
\end{equation}
\begin{equation}
\begin{aligned}
     L_{i2t}^{m}=&-\frac{1}{N}\sum_{i=1}^{N}\log\frac{\exp(s(f_{m,i}^{v},t_{m,i}^{v}))}{\sum\nolimits_{j=1}^{N}\exp(s(f_{m,i}^{v},t_{m,j}^{v}))} \\
     &-\frac{1}{N}\sum_{i=1}^{N}\log\frac{\exp(s(f_{m,i}^{r},t_{m,i}^{r}))}{\sum\nolimits_{j=1}^{N}\exp(s(f_{m,i}^{r},t_{m,j}^{r}))},
\end{aligned}
\end{equation}
\begin{equation}
\begin{aligned}
     L_{t2i}^{m}=&-\frac{1}{N}\sum_{i=1}^{N}\log\frac{\exp(s(t_{m,i}^{v},f_{m,i}^{v}))}{\sum\nolimits_{j=1}^{N}\exp(s(t_{m,i}^{v},f_{m,j}^{v}))} \\
     &-\frac{1}{N}\sum_{i=1}^{N}\log\frac{\exp(s(t_{m,i}^{r},f_{m,i}^{r}))}{\sum\nolimits_{j=1}^{N}\exp(s(t_{m,i}^{r},f_{m,j}^{r}))}.
\end{aligned}
\end{equation}

\noindent(3) Notably, ReID is inherently a single-view retrieval task that measures the similarity between the query and a gallery representation to determine if they belong to the same individual. This implies that multi-view representations are unavailable during inference. To address this, we introduce a knowledge distillation mechanism to propagate multi-view information into the current view representations:
\begin{equation}
\begin{aligned}
L_{kd}&=\frac{1}{N}\sum_{i=1}^{N}\left\|f_{i,m}^{v}-f_{i}^{v}\right\|_{2}^{2}+\frac{1}{N}\sum_{i=1}^{N}\left\|f_{i,m}^{r}-f_{i}^{r}\right\|_{2}^{2}\\ &+\frac{1}{N}\sum_{i=1}^{N}\left\|t_{i,m}^{v}-t_{i}^{v}\right\|_{2}^{2}+\frac{1}{N}\sum_{i=1}^{N}\left\|t_{i,m}^{r}-t_{i}^{r}\right\|_{2}^{2},
\end{aligned}
\end{equation}
where $\left\|\cdot \right\|_{2}^{2}$ indicates Mean Squared Error (MSE) loss. This process enables cross-view semantics compensation on both visual and textual sides.

\subsubsection{Cross-Modality Semantics Purification} One should notice that language descriptions for visible and infrared images often contain inconsistent information, such as color attributes 'blue' versus 'gray', resulting in conflict semantics embedded in paired cross-modality visual representations. To address this, our CMSP constrains the distance between inter-modality image-text pair representations to be close to that of intra-modality image-text pair representations:

\begin{equation}
L_{cmsp}=\frac{1}{N}\sum_{i=1}^{N}(d_{i}^{vv}-d_{i}^{vr})^{2}+\frac{1}{N}\sum_{i=1}^{N}(d_{i}^{rr}-d_{i}^{rv})^{2},
\end{equation}
where $d_{i}^{vv}=\left\|f_{i}^{v}-t_{i}^{v}\right\|_{2}$ and $d_{i}^{vr}=\left\|f_{i}^{v}-t_{i}^{r}\right\|_{2}$ represent Euclidean distances between $f_{i}^{v}$ and $t_{i}^{v}$, and $f_{i}^{v}$ and $t_{i}^{r}$, respectively. Similarly, $d_{i}^{rr}$ and $d_{i}^{rv}$ are defined for the infrared modality. This formula encourages the distances between visual representations of two modalities and the same textual representation to be as equal as possible, thereby eliminating noisy semantics and further enhancing the modality-invariance of visual representations.

\subsection{Training and Inference}
Our EEES is trained in an end-to-end manner, with the total loss can be expressed as:
\begin{equation}
    L=L_{id}+\lambda_{1}L_{wrt}+\lambda_{2}L_{con}+\lambda_{3}L_{kd}+\lambda_{4}L_{cmsp},
\end{equation}
where $L_{con}=L_{con}^{o}+L_{con}^{m}$. The coefficients $\lambda_{1}$, $\lambda_{2}$, $\lambda_{3}$, and $\lambda_{4}$ balance the weights of each loss term. During inference, the language component is not needed, and only single-view visual representations are extracted to measure similarity.

\section{Experiments}
\subsection{Experimental Settings}
\noindent\textbf{Datasets}. \textbf{SYSU-MM01} \cite{vireid} comprises 30,071 visible images captured by 4 RGB cameras and 15,792 infrared images captured by 2 IR cameras. The training set includes 22,258 visible images and 11,909 infrared images of 395 identities. The testing set consists of 3,803 infrared images of 96 identities and either 301 or 3,010 (single-shot or multi-shot) randomly sampled visible images. \textbf{RegDB} \cite{regdb} is a small-scale VIReID dataset containing 4,120 visible images and 4,120 infrared images from 412 pedestrians. Following the protocol \cite{aligngan}, 2,060 visible and 2,060 infrared images of 206 identities are used for training, with the remainder reserved for testing.

\noindent\textbf{Evaluation Metrics}. We assess the retrieval performance using the general indicators named mean Average Precision (mAP) and Cumulative Matching Characteristics (CMC).

\noindent\textbf{Implementation Details}. 
We conduct experiments using the PyTorch library on a single RTX 4090 GPU. Our EEES framework incorporates a training-free LLaVA and fine-tunes CLIP, which includes a visual encoder and a textual encoder, with ResNet50 \cite{resnet50} serving as the backbone for the visual encoder. Following AGW \cite{agw}, we train two parallel first convolutional layers of ResNet50 for each modality while sharing the parameters of the remaining four blocks. During training, we randomly sample 8 identities, each with 4 visible and 4 infrared images. All input images are resized to 288$\times$144 and undergo data augmentation techniques such as random padding, cropping, and flipping. The training process spans 120 epochs, with initial learning rates set to 3e-4 for the visual encoder and 1e-6 for the textual encoder, decaying by 0.1 at the 40th and 70th epochs. Hyper-parameters are set as $\lambda_{1}=0.25$, $\lambda_{2}=0.2$, $\lambda_{3}=0.08$, and $\lambda_{4}=0.01$. Additionally, we set $M=1$, meaning EEES integrates information from two views to construct the multi-view representation.

\subsection{Comparison with State-of-the-Art Methods}
\noindent\textbf{SYSU-MM01}. We evaluate the performance of our EEES on SYSU-MM01 and compare it with state-of-the-art methods. Table \ref{Tab:1} demonstrates that EEES consistently outperforms existing methods across all settings. Specifically, our Rank-1 accuracy and mAP exceed those of the best generative-based method, ACD \cite{acd}, by 3.8\% (2.6\%) and 4.6\% (3.8\%) in the all-search testing mode, and by 7.6\% (4.8\%) and 5.9\% (5.7\%) in the indoor-search mode, respectively. This improvement can be attributed to our method performing modality alignment at the feature level, which circumvents performance limitations imposed by the generated low-quality images. Compared to generative-free methods, our Rank-1 accuracy and mAP surpass ScRL \cite{scrl} by 2.1\% and 3.1\%, and outperform MBCE \cite{mbce} by 3.2\% and 2.6\%. This advantage arises from EEES embedding high-level semantic information into heterogeneous visual contents, facilitating modality alignment. Additionally, our method outperforms CSDN \cite{csdn} across all metrics due to the clear, detailed, and rich semantics learned by EEES, in contrast to the unknown and coarse semantics learned by CSDN.

\noindent\textbf{RegDB}. We conduct further evaluations of EEES on the RegDB dataset, with quantitative results presented in Table 2. Our method achieves superior recognition rates compared to existing generative-based methods. For example, our Rank-1 accuracy outperforms TSME \cite{tsme} by 6.5\%, and our mAP surpasses ACD \cite{acd} by 5.3\% in the visible-to-infrared testing mode. Similarly, our method exhibits significant performance advantages over state-of-the-art generative-free methods. In comparison with CSDN \cite{csdn}, the Rank-1 recognition rate and mAP of EEES are enhanced by 4.8\% (6.0\%) and 3.8\% (5.4\%), respectively. These results comprehensively demonstrate the superiority of our method. 

\begin{table*}[!ht]\small
\centering {\caption{Performance comparison with state-of-the-art methods on SYSU-MM01. '-' denotes that no reported result is available.}\label{Tab:1}
\renewcommand\arraystretch{1.1}
\begin{tabular}{c|c|cc|cc|cc|cc}
\hline
 & & \multicolumn{4}{c|}{All-Search} & \multicolumn{4}{c}{Indoor-Search}\\
  
 \cline{3-10} Methods & Ref & \multicolumn{2}{c|}{Single-Shot} & \multicolumn{2}{c|}{Multi-Shot} & \multicolumn{2}{c|}{Single-Shot} & \multicolumn{2}{c}{Multi-Shot}\\
  
  & & R1 & mAP & R1 & mAP & R1 & mAP & R1 & mAP\\
\hline

  AlignGAN \cite{aligngan} & ICCV'19 & 42.4 & 40.7 & 51.5 & 33.9 & 45.9 & 54.3 & 57.1 & 45.3 \\

  Hi-CMD \cite{hicmd} & CVPR'20 & 34.9 & 35.9 & - & - & - & - & - & - \\

  JSIA \cite{jsia} & AAAI'20 & 38.1 & 36.9 & 45.1 & 29.5 & 43.8 & 52.9 & 52.7 & 42.7 \\

  XIV-ReID \cite{x-modality} & AAAI'20 & 49.9 & 50.7 & - & - & - & - & - & - \\

  TSME \cite{tsme} & TCSVT'22 & 64.2 & 61.2 & 70.3 & 54.3 & 64.8 & 71.5 & 76.8 & 65.0 \\

  ACD \cite{acd} & TIFS'24 & \underline{74.4} & \underline{71.1} & \underline{80.4} & \underline{66.9} & \underline{78.9} & \underline{82.7} & \underline{86.0} & \underline{78.6} \\

  \hline

  NFS \cite{nfs} & CVPR'21 & 56.9 & 55.4 & 63.5 & 48.5 & 62.7 & 69.7 & 70.0 & 61.4 \\
  
   MID \cite{mid} & AAAI'22 & 60.2 & 59.4 & - & - & 64.8 & 70.1 & - & - \\

    MAUM \cite{maum} & CVPR'22 & 71.6 & 68.7 & - & - & 76.9 & 81.9 & - & - \\

    CIFT \cite{cift} & ECCV'22 & 71.7 & 67.6 & 78.0 & 62.4 & 78.6 & 82.1 & 86.9 & 77.0 \\

    MRCN \cite{mrcn} & AAAI'23 & 68.9 & 65.5 & - & - & 76.0 & 79.8 & - & - \\

   CAJ+ \cite{caj+} & TPAMI'23 & 71.4 & 68.1 & - & - & 78.3 & 78.4 & - & - \\

   MBCE \cite{mbce} & AAAI'23 & 74.7 & 72.0 & 78.3 & 65.7 & \underline{83.4} & \underline{86.0} & 88.4 & \underline{80.6} \\

   DEEN \cite{deen} & CVPR'23 & 74.7 & 71.8 & - & - & 80.3 & 83.3 & - & - \\

   SEFL \cite{sefl} & CVPR'23 & 75.1 & 70.1 & - & - & 78.4 & 81.2 & - & - \\

    ScRL \cite{scrl} & arxiv'23 & \underline{76.1} & \underline{72.6} & - & - & 82.4 & 82.2 & - & - \\

   CSMSSF \cite{csmssf} & TMM'24 & 70.5 & 67.4 & - & - & 75.9 & 80.2 & - & - \\

   PMFA \cite{pmfa} & TIM'24 & 74.2 & 70.7 & - & - & 81.1 & 84.1 & - & - \\

   CSDN \cite{csdn} & arxiv'24 & 75.2 & 71.8 & \underline{80.6} & \underline{66.3} & 82.0 & 85.0 & \underline{88.5} & 80.4 \\

  \hline
   \textbf{Ours (EEES)} & - & \textbf{78.2} & \textbf{75.7} & \textbf{83.0} & \textbf{70.7} & \textbf{86.5} & \textbf{88.6} & \textbf{90.8} & \textbf{84.3} \\
   \hline
\end{tabular}}
\end{table*}

\begin{table}[!ht]\small
\centering {\caption{Performance comparison on RegDB.}\label{Tab:2}
\renewcommand\arraystretch{1.1}
\begin{tabular}{c|cc|cc}
\hline
 \multirow{2}*{Methods} & \multicolumn{2}{c|}{Visible to Infrared} & \multicolumn{2}{c}{Infrared to Visible}\\

  & R1 & mAP & R1 & mAP \\
 \hline

 AlighGAN & 56.3 & 53.4 & 57.9 & 53.6\\

 Hi-CMD & 70.9 & 66.0 & - & - \\

 JSIA & 48.1 & 48.9 & 48.5 & 49.3 \\

 XIV-ReID & - & - & 62.2 & 60.1\\

 GECNet & 82.3 & 78.4 & 78.9 & 75.5\\

 TSME & \underline{87.3} & 76.9 & 86.4 & 75.7 \\

  ACD & 84.7 & \underline{83.2} & \underline{87.1} & \underline{84.7} \\

 \hline

 NFS & 80.5 & 72.1 & 77.9 & 69.7 \\

 MID & 87.4 & 84.8 & 84.2 & 81.4\\

 MAUM & 87.8 & 85.0 & 86.9 & 84.3\\

 CIFT & 92.1 & 86.9 & 90.1 & 84.8 \\

 MRCN & 91.4 & 84.6 & 88.3 & 81.9 \\

 CAJ+ & 85.6 & 79.7 & 84.8 & 78.5 \\

 MBCE & \underline{93.1} & \underline{88.3} & \underline{93.4} & \underline{87.9}\\

 DEEN & 91.1 & 85.1 & 89.5 & 83.4\\

 SEFL & 91.0 & 85.2 & 92.1 & 86.5 \\

 ScRL & 92.4 & 86.7 & 91.8 & 85.3 \\

 CSMSSF & 85.3 & 76.3 & 83.8 & 75.1\\

 PMFA & 92.3 & 84.7 & 91.1 & 83.5\\

 CSDN & 89.0 & 84.7 & 88.2 & 82.8 \\

\hline
\textbf{Ours (EEES)} & \textbf{93.8} & \textbf{88.5} & \textbf{94.2} & \textbf{88.2} \\

  \hline
\end{tabular}}
\end{table}

\begin{table}[!ht]\small
\centering {\caption{Ablation studies of our EEES.}\label{Tab:3}
\renewcommand\arraystretch{1.2}
\begin{tabular}{c|c|ccc|cc}
 \hline
  \multicolumn{2}{c|}{Methods} & ESE & CVSC & CMSP & R1 & mAP \\
\hline

  \multicolumn{2}{c|}{Baseline} &  & & & 71.6 & 68.0 \\

  \multicolumn{2}{c|}{ISE} &  & & & 74.1 & 71.8 \\
  \hline

  \multirow{4}*{EEES} & \ 1 & \checkmark & & & 75.4 & 72.5 \\

  & \ 2 & \checkmark & \checkmark & & 76.7 & 74.4 \\

  & \ 3 & \checkmark & & \checkmark & 76.2 & 74.2 \\

  & \ 4 & \checkmark & \checkmark & \checkmark & \bf{78.2} & \bf{75.7} \\

  \hline
\end{tabular}}
\end{table}

\subsection{Ablation Studies}

We evaluate the effectiveness of each component in our EEES framework, with the results presented in Table \ref{Tab:3}. The Baseline represents addressing VIReID solely through a vision-centric approach, while ISE denotes implicit semantics embedding using the prompt learner.

\noindent\textbf{Effectiveness of ESE}. ESE replaces the prompt learner in the ISE with language descriptions generated using LLaVA, capturing explicit pedestrian semantics. Consequently, it improves the Rank-1 and mAP by  1.3\% and 0.7\% compared to ISE. This improvement is attributed to the clarity and detail of our explicit semantics, validating both the rationale behind our motivation and the effectiveness of our technology.

\noindent\textbf{Effectiveness of CVSC}. CVSC integrates cross-view information into the single-view representation to enrich pedestrian semantics. When CVSC is equipped with ESE, the Rank-1 and mAP accuracy rates are improved by 1.3\% and 1.9\%, respectively. This confirms the comprehensiveness of multi-view information and the effectiveness of CVSC in compensating for cross-view semantics.

\noindent\textbf{Effectiveness of CMSP}. CMSP constrains that the distance between inter-modality image-text pair representations is equal to that between intra-modality image-text pair representations, thereby preventing the embedding of noisy semantics. It improves Rank-1 and mAP by 0.8\% and 1.7\%, respectively, when added to ESE. Moreover, when combined with both ESE and CVSC, performance reaches 78.2\% and 75.7\%, respectively, demonstrating its effectiveness in eliminating noisy semantics and enhancing the modality-invariance of heterogeneous visual representations.

\section{Further Discussion}

\subsection{Parameters Analysis}
\begin{figure}[t!]
  \centering
{\includegraphics[height=1.1in,width=1.6in,angle=0]{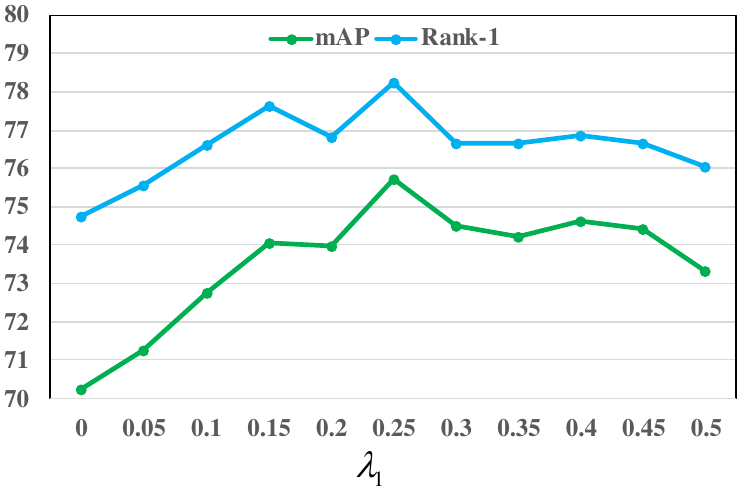}} {\includegraphics[height=1.1in,width=1.6in,angle=0]{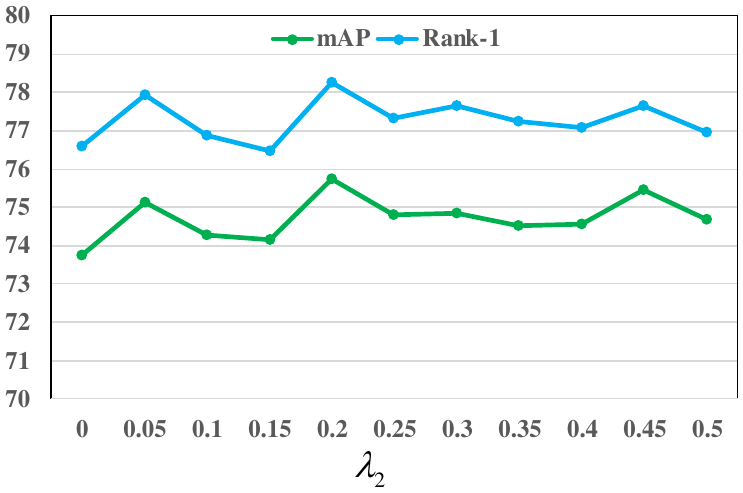}}
{\includegraphics[height=1.1in,width=1.6in,angle=0]{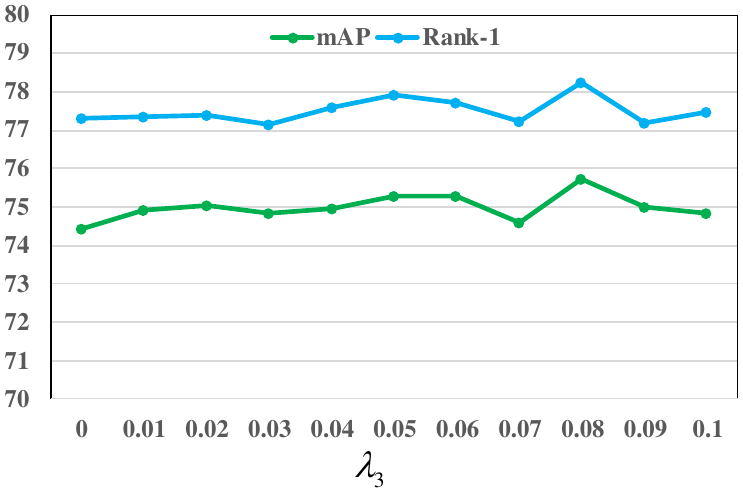}} {\includegraphics[height=1.1in,width=1.6in,angle=0]{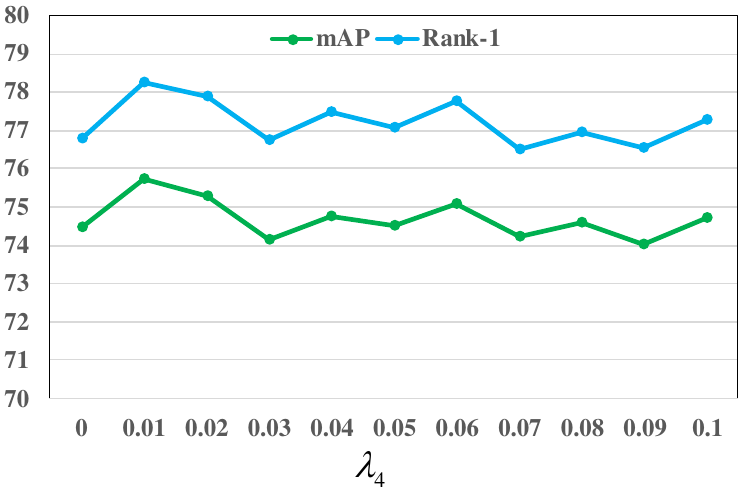}}\\
  \centering\caption{Parameters analysis of $\lambda_{1}$, $\lambda_{2}$, $\lambda_{3}$, and $\lambda_{4}$.}
  \label{Fig:3}
\end{figure}

The hyper-parameters $\lambda_{1}$, $\lambda_{2}$, $\lambda_{3}$, and $\lambda_{4}$ regulate the relative importance of each loss term in our EEES framework. Figure \ref{Fig:3} demonstrates that the optimal values for these hyper-parameters are 0.25, 0.2, 0.08, and 0.01, respectively. Moreover, setting any of these values to 0 results in decreased performance, affirming the rationality and effectiveness of each proposed loss term.

\subsection{Number of Cross-View Representations}
The proposed CVSC compensates semantics from $M$ cross-view representations into the single-view one. Table \ref{Tab:4} shows the effects of varying $M$ values on performance. When $M=1$, both Rank-1 accuracy and mAP reach the peak, indicating that integrating information from two views comprehensively characterizes pedestrians. Conversely, performance declines when $M=0$ and $M>1$. The former underscores the rationality and effectiveness of our CVSC, while the latter may result from increased pedestrian-independent view noise, such as background information.

\begin{table}[!ht]\small
\centering {\caption{Effects of the number of cross-view representations.}\label{Tab:4}
\renewcommand\arraystretch{1.1}
\begin{tabular}{c|cc|cc}
\hline

 \cline{2-5} \multirow{2}{*}{Number} & \multicolumn{2}{c|}{Single-Shot} & \multicolumn{2}{c}{Multi-Shot}\\

 & R1 & mAP & R1 & mAP\\

 \hline

 0 & 76.1(86.3) & 74.2(83.7) & 81.8(89.4) & 67.5(81.8) \\

 1 & \textbf{78.2}(\textbf{86.5}) & \textbf{75.7}(\textbf{88.6}) & \textbf{83.0}(\textbf{90.8}) & \textbf{70.7}(\textbf{84.3})\\

 2 & 77.1(86.5) & 74.6(88.4) & 82.8(90.1) & 70.0(83.9) \\

 3 & 77.5(86.3) & 74.9(88.5) & 83.0(90.3) & 70.1(84.1)\\

 \hline
\end{tabular}}
\end{table}

\subsection{Visualization}
\begin{figure}[th!]
\centering
{\includegraphics[height=3.5in,width=3.3in,angle=0]{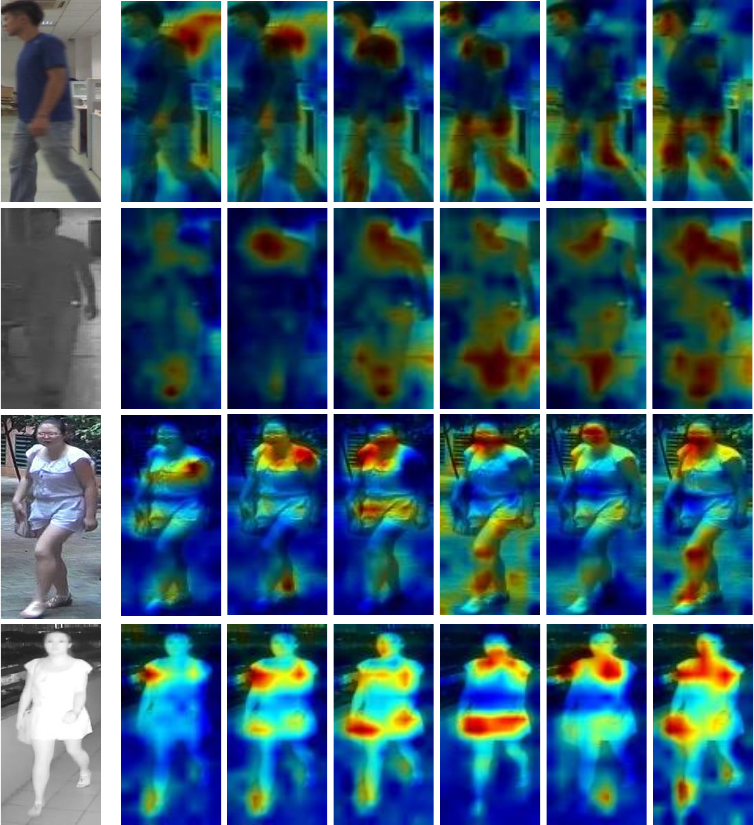}}
\caption{Visualization of spatial discriminative regions. From left to right, the images are arranged as follows: the original image, followed by heatmaps of Baseline, ISE, ESE, ESE+CVSC, ESE+CMSP, and EEES.}
\label{Fig:4}
\end{figure}

Our EEES framework learns visual representations associated with high-quality semantics through three key aspects: embedding explicit semantics, compensating for cross-view semantics, and eliminating noisy semantics. Figure \ref{Fig:4} shows spatial discriminative regions of interest identified by the model using Class Activation Maps (CAMs) \cite{cams}. As we can see, ESE directs the model to focus on more identity-related discriminative features compared with the Baseline and ISE. CVSC broadens the areas identified by ESE, while CMSP ensures the model emphasizes clues such as the face and legs rather than clothing. Overall, each module effectively fulfills its intended purpose.

\subsection{Limitations}
The large language-vision generation model provides language descriptions with explicit pedestrian semantics. However, it may produce incorrect descriptions, especially for low-resolution infrared images, as it is not pre-trained on large-scale pedestrian image-text pairs and has not encountered infrared images. Additionally, we observed that performance decreased when the number of cross-views involved in information integration increased, likely due to the enhancement of view noise. This motivates us to design better information integration approaches to compensate for cross-view semantics into single-view representation in the future.

\section{Conclusion}
In this paper, we propose a novel Embedding and Enriching Explicit Semantics (EEES) framework to embed high-quality semantics into heterogeneous visual representations, effectively alleviating the modality discrepancy in VIReID. Our EEES is the first to connect with multiple mainstream large language-vision models, automatically supplementing language descriptions to capture explicit pedestrian semantics. Our EEES also considers the complementarity of multi-view information, exploring the many-to-many correspondences of image-text pairs to compensate for cross-view semantics. Furthermore, our EEES constrains the distance consistency of image-text pairs across modalities, eliminating conflicting semantics in heterogeneous visual representations. Experimental results on two datasets demonstrate the superiority and effectiveness of our proposed method.

\bibliography{aaai25}

\end{document}